\pdfoutput=1
\documentclass[11pt]{article}

\usepackage[]{acl}

\usepackage{times}
\usepackage{latexsym}

\usepackage[T1]{fontenc}
\usepackage[utf8]{inputenc}

\usepackage{microtype}

\usepackage{algorithm}
\usepackage{algorithmic}
\usepackage{amssymb,amsmath,amsthm}
\usepackage{xcolor}
\usepackage{graphicx}
\usepackage[utf8]{inputenc}
\usepackage[english]{babel}
\usepackage{multirow}
\usepackage{booktabs}

\newtheorem{remark}{Remark}
\newtheorem{theorem}{Theorem}
\usepackage{algorithm}

\DeclareMathOperator*{\argmin}{arg\,min}
\usepackage{amssymb}
\usepackage{mathtools}

\title{LABO: Towards Learning Optimal Label Regularization via Bi-level Optimization}
\author{Peng Lu$^{1}$, Ahmad Rashid$^{2, 3}$, Ivan Kobyzev$^2$ \\ \textbf{Mehdi Rezagholizadeh$^2$,  Philippe Langlais$^1$}  \\
$^1$ Department of Computer Science and Operations Research, Universit\'e de Montr\'eal\\
 $^2$ Huawei Noah’s Ark Lab, Canada\\
$^3$ Department of Statistics and Actuarial Science, University of Waterloo \\
{peng.lu@umontreal.ca}}

\begin{document}
\maketitle
\begin{abstract}
Regularization techniques are crucial to improving the generalization performance and training efficiency of deep neural networks. Many deep learning algorithms rely on weight decay, dropout, batch/layer normalization to converge faster and generalize. Label Smoothing (LS) is another simple, versatile and efficient regularization which can be applied to various supervised classification tasks. Conventional LS, however, regardless of the training instance assumes that each non-target class is equally likely.
In this work, we present a general framework for training with label regularization, which includes conventional LS but can also model instance-specific variants. Based on this formulation, we propose an efficient way of learning LAbel regularization by devising a Bi-level Optimization (LABO) problem. We derive a deterministic and interpretable solution of the inner loop as the optimal label smoothing without the need to store the parameters or the output of a trained model. 
Finally, we conduct extensive experiments and demonstrate our LABO consistently yields improvement over conventional label regularization on various fields, including seven machine translation and three image classification tasks across various neural network architectures while maintaining training efficiency.
\end{abstract}

\section{Introduction}

Deep neural networks (DNNs)  form the backbone of current state-of-the-art algorithms in various fields including natural language processing~\citep{DBLP:conf/nips/VaswaniSPUJGKP17}, computer vision~\citep{DBLP:conf/cvpr/HeZRS16, DBLP:conf/iclr/DosovitskiyB0WZ21} and speech recognition~\citep{DBLP:conf/interspeech/SchneiderBCA19, DBLP:journals/corr/abs-2110-13900}. However, heavily overparameterized models may incur overfitting and suffer from poor generalizations~\citep{DBLP:books/daglib/0040158}. To address the issue, many regularization techniques have been developed in the literature: weight decay which constrains the optimization space~\citep{DBLP:conf/nips/KroghH91}, batch or layer normalization which speeds up the training of feed-forward NNs~\citep{DBLP:conf/icml/IoffeS15, DBLP:journals/corr/BaKH16}, and dropout which implicitly approximates the effect of averaging the predictions of all sparse subnetworks networks~\citep{DBLP:journals/jmlr/SrivastavaHKSS14}. 
% Besides these regularization techniques, 
Label smoothing (LS) is another simple regularization technique; it is widely applied to many applications including image classification ~\citep{szegedy2016rethinking} and token-level sequence generation~\cite{DBLP:conf/iclr/PereyraTCKH17} for enhancing the generalization,  without suffering additional computational costs. It encourages a model to treat each non-target class as equally likely for classification by using a uniform distribution to smooth one-hot labels~\citep{DBLP:conf/cvpr/HeZ0ZXL19, DBLP:conf/nips/VaswaniSPUJGKP17}. Although combining the uniform distribution with the original one-hot label is beneficial for regularization, conventional LS does not take into account the true relations between different label categories. More specifically, for token-level generation, uniformly allocating the probability mass on non-target words disregards the semantic relationship between non-target words and the context. On the other hand, the probability of the target is predefined and unchanged. However, the distribution of natural language exhibits remarkable variations in the per-token perplexity~\cite{DBLP:conf/iclr/HoltzmanBDFC20}, which encourages us to adapt
corresponding target probabilities for different contexts.   

One of the instance-dependent techniques of learning the relation between different target categories is Knowledge Distillation (KD)~\citep{hinton2015distilling,bucilua2006model}, which is a popular technique of transfer learning utilizing knowledge from related tasks or teacher models~\cite{mtl_book, tl_survey, sc-lstm}. It is widely applied for model compression and ensembling across applications ranging from computer vision~\citep{He2019BagOT, xu2020normalized, rw-kd} to natural language processing~\citep{jiao2020tinybert,sanh2019distilbert}. However, KD requires training a separate model as a teacher for every new task. Besides, it either introduces an extra inference pass to get the teacher's prediction for each instance during the training or requires saving the teacher's prediction for all training samples to avoid the extra forward pass of teacher models. This greatly increases the time and space complexity in practice. Especially for token-level generation tasks, e.g. machine translation, to save all the output probabilities of teacher models costs $O(N L  V)$ space, where $N$ is the number of sequences, $L$ is the averaged length of all sequences and $V$ is the vocabulary size.
Besides the empirical success of KD, it is unclear how student networks benefit from these smoothed labels. A series of investigations have looked at regularization and have demonstrated that the success of both KD and label smoothing is due to a similar regularization effect of smoothed targets~\citep{yuan2020revisiting, DBLP:conf/nips/ZhangS20}. Based on this finding and the low training overhead of LS, there is a significant interest, in the community, in algorithms that can enhance conventional LS. ~\cite{DBLP:conf/nips/ZhangS20}  demonstrates the importance of an instance specific LS regularization. They demonstrate better performance compared to LS, but use a trained model to infer prior knowledge on the label space and thereby sacrifice some of the efficiency of LS.

In this work, we first revisit the conventional LS and generalize it to an instance-dependent label regularization framework with a constraint on over-confidence.
Within this framework, we demonstrate that both LS and KD can be interpreted as instances of a smoothing distribution with a confidence penalty. Finally, we propose to learn the optimal smoothing function along with the model training, by devising a bi-level optimization problem. We solve the inner problem by giving a deterministic and interpretable solution and applying gradient-based optimization to solve the outer optimization. 

Our contributions can be summarized as follows:

\begin{itemize}
\item We explicitly formulate a unified label regularization framework, in which the regularization distribution can be efficiently learnt along with model training by considering it as a bi-level optimization problem.

\item We derive a closed-form solution to solve the inner loop of the bi-level optimization, which not only improves efficiency but makes the algorithm interpretable.
\item We conducted extensive experiments on Machine Translation, IWSLT'14 (DE-EN, EN-DE, EN-FR, FR-EN), WMT'14 (EN-DE, DE-EN), IWLST'17 (\{DE,FR\}-EN), image classification (CIFAR10, CIFAR100 and ImageNet) and show that our method outperforms label smoothing consistently while maintaining the training efficiency.

\end{itemize}
\section{Background}
We will provide a brief overview of existing label regularization methods.

\paragraph{Label Smoothing.}
LS is a regularization technique to improve the generalization performance of neural networks by preventing the model from predicting the training examples overconfidently.
It smoothes a one-hot target label with the uniform distribution $U(\cdot) = \frac{1}{K}$, where $K$ is the number of classes. As a result, LS training is equivalent to training with a smoothed label $\hat{P}_{U}$, where:

\begin{align}
\label{eq:ls_u}
    \hat{P}_{U}(j|x) = 
    \begin{cases}
    1 - \alpha + \alpha U(j), & j=k\\
    \alpha U(j), & j \neq k
    \end{cases},
\end{align}
and $k$ is the ground-truth class.  We note from Eq.~\ref{eq:ls_u}  that a higher $\alpha$ can lead to a smoother label distribution and a lower $\alpha$ to a \textit{peakier} label.

\paragraph{Confidence Penalty.} Another important technique similar to label smoothing is the confidence penalty (CP)~\cite{DBLP:conf/iclr/PereyraTCKH17}, in which a regularization term $\mathbb{H}(p_\theta(x))$  is introduced into the objective function to directly punish the over-confidence of predictions:
\begin{align}
    \mathbb{H}(q(x),p_\theta(x)) - \beta \mathbb{H}(p_\theta(x))
\end{align}
where $q(x)$ is the one-hot label, $p_\theta(x)$ is the output distribution of models, $\mathbb{H}(q(x),p_\theta(x))$ is the cross-entropy loss between the labels and the student's output, $\mathbb{H}(p_\theta(x))$ is the entropy of the output and $\beta$ controls the strength of the confidence penalty.

\paragraph{Knowledge Distillation.} Given access to a trained teacher model, assume that we want to train a student. Denote by $P_T(x)$ and $p_\theta(x)$  the teacher's and student's predictions respectively. For a classification problem, the total KD loss is defined as:
\begin{align}
    (1-\alpha)\mathbb{H}(q(x),p_\theta(x)) +\alpha \mathbb{KL}(P_T(x),p_\theta(x)),
\end{align}
where $\mathbb{KL}$ is the Kullback–Leibler (KL) divergence and $\alpha$ is the scaling parameter. Note that we assume a temperature of 1 and omit it without loss of generality. KD training is equivalent to training with a smoothed label $\hat{P}(x)$, where:

\begin{align}
    \hat{P}_{T}(j|x) = 
    \begin{cases}
    1 - \alpha + \alpha P_T(j|x), & j=k\\
    \alpha P_T(j|x), & j \neq k
    \end{cases},
\end{align}

where $k$ is the ground-truth class.

Both LS and KD incorporate training a model with a smoothed label. KD tends to perform better as the teacher's predictions over non-target classes can  capture the similarities between different classes~\citep{DBLP:conf/nips/MullerKH19, shen2021label}. 

However, LS and CP techniques are more efficient for training, since they do not require training a separate teacher network for every new task. 

\begin{figure*}[t!]
\begin{center}
\includegraphics[width=0.99\textwidth]{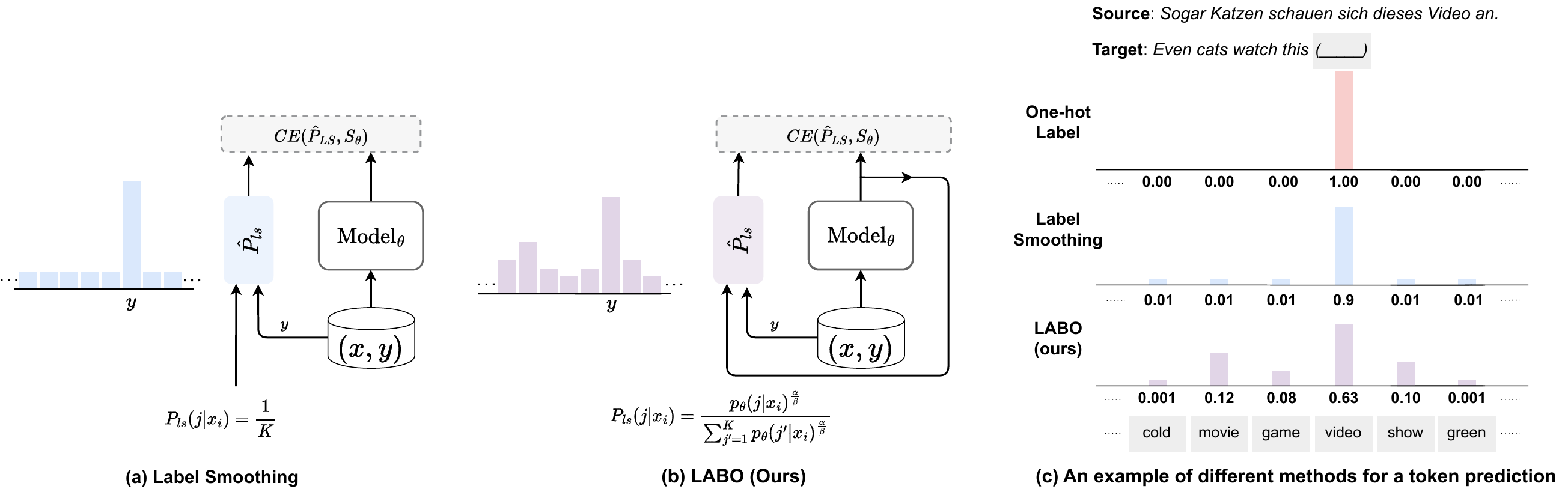} % Reduce the figure size so that it is slightly narrower than the column.
\caption{Comparison of Label Smoothing and LABO. a) shows Label Smoothing. b) shows LABO. c) gives an illustration of a one-hot target, label smoothing and our LABO method given a translation pair sampled from the IWSLT'14 (DE-EN) dataset. It shows a part of the probability vector when predicting the next ground-truth word "\textit{video}". Our method assigns larger probabilities to the context-relevant tokens compared to label smoothing.} 
\label{fig2}
\end{center}
\end{figure*}

\section{Methodology}
In this section, we interpret our optimal label regularization from the bi-level optimization perspective.
First, we provide a close look at the conventional uniform label regularization and show the generalized framework bridges the objectives of conventional LS and KD methods. Then, we introduce a closed-form solution to our optimal instance-dependent label smoothing and describe an online implementation under this formulation.

\subsection{Generalized Label Regularization:  A Close Look}
Suppose that we have a $K$-class classification task to be learned by a neural network, $S_\theta(\cdot)$.  Given a training set of $\{x_i, y_i\}_{i=1}^N$ samples where $x_i$ is a data sample and $y_i$ is the ground-truth label,  the model $S_\theta(\cdot)$ outputs the probability $p_\theta(j|x_i)$ of each class $j \in \{1, \dots, K\}$:
\begin{align}
   p_\theta(j|x_i) = \frac{\exp(z_{i,j})}{\sum_{j'=1}^K\exp(z_{i,j'})},
\end{align}
where $z_{i,j} = [S_\theta(x_i)]_j$ is the logit for $j$-th class of input $x_i$. 

A general label smoothing can be formally written as:
\begin{align}
\label{def:p_hat}
    \hat{P}_{ls}(j|x_i) = 
    \begin{cases}
    1 - \alpha + \alpha\cdot P_{ls}(j|x_i), & j=k\\
    \alpha\cdot P_{ls}(j|x_i), & j \neq k
    \end{cases},
\end{align}

where $\hat{P}_{ls}$ is a smoothed label, $P_{ls}$ is a smoothing distribution, $k$ is the ground-truth class of $x_i$, and $\alpha$ is a hyperparameter that determines the amount of smoothing. If $\alpha = 0$, we obtain the original one-hot encoded label.
For the original label smoothing method, the probability $P_{ls}(\cdot |x)$ is independent on the sample $x$ and is taken to be the uniform distribution ${P}_{ls}(j) = U(j)$, However, the training can benefit from instance-dependent regularization.~\citep{yuan2020revisiting, DBLP:conf/nips/ZhangS20}. In this work, we consider the general form of LS $\hat{P}_{ls}(\cdot \vert x_i)$ which is instance-dependent and not necessarily uniform.  

Let us consider the Cross Entropy (CE) loss with the smoothed labels:
\begin{align}
    \mathcal{L}_\theta(P_{ls})
    % &= \frac{1}{N}\sum_{i=1}^N  (1-\alpha)\mathcal {L}_{CE} + \alpha\mathcal{L}_{ls} \nonumber \\
    % &=  \frac{1}{N}\sum_{i=1}^N \left[-(1-\alpha)\log p(k|x_i) -  \alpha\sum_{j=1}^K\frac{1}{K}\log p(j|x_i) \right] \nonumber \\
    % &=  \frac{1}{N}\sum_{i=1}^N \left[-(1-\alpha + \frac{\alpha}{K})\log p(k|x_i) -  \sum_{j=1, j\neq k}^K\frac{\alpha}{K}\log p(j|x_i) \right] \nonumber \\
    &= \frac{1}{N}\sum_{i=1}^N \left[ -\sum_{j=1}^K \hat{P}_{ls}(j|x_i)\log p_\theta(j|x_i) \right] \nonumber\\
    &= \frac{1}{N}\sum_{i=1}^N \mathbb{E}_{\hat{P}_{ls}}\left[ - \log p_\theta(j|x_i)  \right].
\label{eq:ls}
\end{align}

Note that computing the weighted sum of the negative log-likelihood of each probability of label can be viewed as taking expectation of the negative log-likelihood over label space under a certain distribution $\hat{P}_{ls}$.
We modify this loss by adding a KL divergence term $\mathbb{KL}(P_{ls}(\cdot|x_i) \Vert U(\cdot))$ into Eq.~\ref{eq:ls} which encourages the sample-wise smoothing distribution $P_{ls}(\cdot|x_i) $ to be close to the uniform distribution to handle the over-confidence. 
\begin{align}
\label{eq:pseudo-kd}
\begin{split}
     \mathcal{R}_\theta(P_{ls}) = \frac{1}{N}\sum_{i=1}^N  \Big[ &\mathbb{E}_{\hat{P}_{ls}}\left[ -\log p_\theta(j|x_i)  \right] + \\  &\beta\mathbb{KL}(P_{ls}(\cdot|x_i) \Vert U(\cdot))\Big],
\end{split}
\end{align}

where $\beta$ is a hyper-parameter. The instance-dependent smoothing $P_{ls}(\cdot|x_i) $ can be viewed as the prior knowledge over the label space for a sample $x_i$. This KL divergence term can be understood as a measure of the `smoothness' or `overconfidence' of  each training label. Specifically for token-level generation tasks, the over-confidence of model results in the output of repetitive or most frequent but irrelevant text, which is detrimental to the generalization of the model~\cite{DBLP:conf/interspeech/ChorowskiJ17, DBLP:conf/iclr/HoltzmanBDFC20, DBLP:conf/acl/MeisterSC20}. We choose a Uniform distribution to constrain $P_{ls}$ in the KL term because we would like $P_{ls}$ to contain more information on non-target classes. It plays a role similar to the temperature in KD and  controls the sharpness of the distribution $P_{ls}$. On the one hand side, in case of KD, it regulates the amount of prior knowledge we inject in the smoothed label. On the other hand side, it reduces the overconfidence of the trained network by increasing the entropy of  smoothed labels (a phenomenon studied by \citet{DBLP:conf/nips/ZhangS20}).

Next, we show the post-hoc interpretability of this formulation. The following two Remarks discuss the relationship of this objective with conventional LS, confidence penalty and KD.
\begin{remark}
When $P_{ls}(\cdot \vert x_i)$ is taken to be the uniform distribution $U(\cdot)$ for any $x_i$,  the objective in Eq.~\ref{eq:pseudo-kd} reduces back to the one in Eq.~\ref{eq:ls} since $\mathbb{KL}\big(U (\cdot) \Vert U (\cdot )\big) = 0$.
\end{remark}

\begin{remark}
This framework can include KD with temperature $\tau = 1$ as a special case.
Suppose $P_T(\cdot|x_i)$ to be the output probability of a teacher model $T(\cdot)$ for any $x_i$, the objective of KD can be rewritten as an expectation of negative log-likelihood w.r.t. a transformed distribution $\hat{P}_T$ plus a KL term between the teacher output and uniform distribution. 
\begin{align}
\label{eq:kd_kl}
\begin{split}
    \mathcal{R}_\theta(P_T)&= \frac{1}{N}\sum_{i=1}^N \Big[ \mathbb{E}_{\hat{P}_T}\left[ - \log p_\theta(j|x_i)  \right] + \\  
    & \alpha \mathbb{KL}(P_T(\cdot|x_i)\Vert U(\cdot))\Big]+ \Theta(\log K),
\end{split}
\end{align}
where $\hat{P}_T(j\vert x_i) = (1 - \alpha + \alpha  \cdot P_T(k\vert x_i))$ for $j = k$ the ground truth and $\hat{P}_T(j\vert x_i) = \alpha  \cdot P_T(j\vert x_i)$ for $j \neq k$. This objective function Eq.~\ref{eq:pseudo-kd} bridges the objective of label smoothing and knowledge distillation, since it is easy to convert our objective to KD or LS by replacing the $P_{ls}(\cdot \vert x_i)$ in Eq.~\ref{eq:pseudo-kd} with $P_T(\cdot \vert x_i)$ or $U(\cdot)$, respectively. There is an inherent relation among these methods. KD and LS, in particular, can be interpreted as instances of our method with a fixed smoothing distribution, which is consistent with recent work~\cite{yuan2020revisiting, DBLP:conf/nips/ZhangS20}.
\end{remark}

\subsection{Learning Label regularization via Bi-level  Optimization}
The choice of the smoothing distribution $P_{ls}$ determines the label regularization method. As described above, in KD the smoothing distribution is an output of the pre-trained teacher model, and in LS it is  a uniform distribution. 
Generally speaking,  the optimal smoothing distribution is unknown, and ideally, we would like to learn the optimal $P_{ls}$.

In this regard, we set up the following two-stage optimization problem: 
\begin{align}
\begin{split}
    & \min_\theta \mathcal{R}(P^*_{ls}(\theta), \theta), \\ \text{subject to} &\ P^*_{ls}(\theta) = \argmin_{P_{ls}} \mathcal{R}_\theta(P_{ls}).
\end{split}
\label{eq:dual}
\end{align}
This optimization setting, also called bilevel optimization \citep{colson2007}, is strongly NP-hard \citep{jeroslow1985} so getting an exact solution is difficult. To solve this problem in our case, we first prove that the inner loop is a convex optimization problem by computing the Hessian matrix $\mathbf{H}_i$ of $\mathcal{R}_\theta(P_{ls})$ w.r.t. $P_{ls}(\cdot\vert x_i)$ for each training instance $x_i$.
\begin{align}
    \mathbf{H}_i=\text{diag}(\frac{\beta}{P_{ls}(1)}, \frac{\beta}{P_{ls}(2)}, \cdots, \frac{\beta}{P_{ls}(K)})
\end{align}
When $\beta$ is greater than zero, the Hessian is positive definite.
Therefore, for the inner optimization loop we can derive the closed-form solution by using a Lagrangian multiplier.
The following Theorem~\ref{theorem1} gives the explicit formulation of $\mathcal{R}_\theta(P_{ls}^*)$. For the details of this derivation please refer to Appendix A. 

\begin{theorem} The solution to the inner loop of optimization in Eq.~\ref{eq:dual} is given by:
\label{theorem1}
\begin{equation}
    P^{*}_{ls}(j|x_i) = \frac{ p_\theta(j|x_i)^\frac{ \alpha}{\beta}}{\sum_{j'=1}^K  p_\theta(j'|x_i)^\frac{\alpha}{\beta}}, 
\end{equation}
where 
$p_\theta(j|x_i)$ is the output probability of $j$-th class of model $S_{\theta}(\cdot)$, $\alpha$ is the smoothing coefficient defined in Eq.~\ref{def:p_hat}, and $\beta$ is  defined in Eq.~\ref{eq:pseudo-kd}. 
\end{theorem}

As a result, we reduced the two-stage optimization problem in Eq.~\ref{eq:dual} to a regular single-stage minimization:
\begin{equation}
    \min_\theta \frac{1}{N}\sum_{i=1}^N\mathcal{R}_\theta(P^*_{ls}(\cdot | x_i)),  
\end{equation}
where
\begin{align}
\label{eq:theorem}
\begin{split}
        \mathcal{R}_\theta(P_{ls}^*(\cdot | x_i)) =  \sum_{j=1}^K\big[ -\hat{P}_{ls}^*(j|x_i)\log p_\theta(j|x_i) \\
    + \beta P^{*}_{ls}(j|x_i) \log(K\cdot P^{*}_{ls}(j|x_i))\big],
\end{split}
\end{align}
$P^{*}_{ls}(j|x_i)$ is given in Theorem~\ref{theorem1}, and $\hat{P}^{*}_{ls}(j|x_i)$ is defined in Eq.~\ref{def:p_hat}.
Note that the solution $P^*_{ls}$ is deterministic and interpretable. Moreover, the two remarks below demonstrate the relation of our LABO with  LS and KD methods.

\begin{remark}
When $\beta$ is extremely large, $P^*_{ls}$ will be close to the Uniform distribution, and our objective function will be equivalent to optimizing the CE loss with a Uniform LS regularizer. 
\end{remark}
\begin{remark}
There is an intrinsic connection between the $P_{ls}^*$ distribution and generating softmax outputs with a temperature factor. Specifically, when $\beta = \alpha \cdot \tau$, we could have 
\begin{align}
    P^*_{ls}(j\vert x_i) = \frac{\exp({\frac{z_{i,j}}{\tau}})}{\sum_{j'=1}^K \exp({\frac{z_{i,j'}}{\tau}})}.
    \label{eq:temp}
\end{align}
The smoothing distribution in this case becomes the temperature smoothed output of the model, which is similar as the smoothed targets used in KD methods\footnote{The derivation can be found in Appendix B.}. 
\end{remark}

To summarize, our method can be expressed as an alternating two-stage process. We generate optimal smoothed labels $P^*_{ls}$ using Theorem~\ref{theorem1} in the first stage. Then, in the second phase, we fix the $P^*_{ls}$ to compute loss $\mathcal{R}_\theta(P^*_{ls})$ and update the model parameters. 

\subsection{Implementation Details}
\paragraph{Two-stage training.} Our solution provides the closed-form answer to the inner loop optimization (1st-stage) and for the outer loop (2nd-stage) the model $f(\theta)$ is updated by gradient-descent. LABO conducts a one-step update for the 2nd stage, namely, for each training step, we compute the optimal smoothing $P^*_{\text{ls}}$ and update the model, which eliminates the need for additional memory or storage for the parameters or outputs of a prior, trained model. The training process is shown in Algorithm~\ref{algo}.

\paragraph{Adaptive $\alpha$.} 
The value of $\alpha$ determines the probability mass to smooth. To get rid of hyper-parameter searching, we provide an instance-specific $\alpha$ as a function of the entropy of the output. 
\begin{align}
      \alpha = \frac{\mathbb{H}(U)-\rho\mathbb{H}(P_{\theta})}{\mathbb{H}(U)},
\end{align}
where $\rho \in [0.5, 1]$, in our experiments, we use $\rho=0.5$. In the experiments, we fix the ratio of $\frac{\beta}{\alpha}$ as a hyper-parameter, so the value of $\beta$ will change accordingly.

\paragraph{Hypergradient} In the outer loop, the derivative of loss $\mathcal{R}(P^*_{ls}(\theta), \theta)$ w.r.t. $\theta$ consists of two components:
\begin{align}
      \nabla_{P^*_{ls}}\mathcal{R}\nabla_{\theta}P^*_{ls} + \nabla_{\theta}\mathcal{R}
\end{align}

because $P^*_{ls}$ is the global solution of objective $\mathcal{R}$ in the inner loop, $\nabla_{P^*_{ls}}\mathcal{R}$ equals zero. Therefore, the hypergradient equals zero, we neglect this component in computation for efficiency.

\begin{algorithm}[htbp]
 \textbf{Input:} Training set $D_{\text{train}}$, batch size $n$, number of steps $T$, learning rate $\eta$, $\hat{P}_{ls}$ warm-up steps $T_w$;
  \begin{algorithmic}[1]
  
    \FOR {$i\leftarrow 1$ \textbf{to} $T$}
    \STATE  Sample a mini-batch $S = \{ (x_i, y_i)\}|_{i=1}^{n}$ from $D_{\text{train}}$;
    \IF{$i < T_w$}
    \STATE Compute the $\hat{P}_{ls}$ with Uniform Distribution for the mini-batch data;
    
    \ELSE
    \STATE Compute the $\hat{P}_{ls}$ according to $\hat{P}_{ls}^*$ solution for the mini-batch data; 
    \ENDIF
    \STATE Update $\theta_{t+1} = \theta_{t} - \eta \nabla_\theta R_{\theta}(\hat{P}_{ls}^*, S)$; 
    % \ENDFOR
    \ENDFOR
    \end{algorithmic}
  \caption{LABO: Two-stage training}\label{algo}  
\end{algorithm}

\begin{table*}[t!]
    % \centering
    \caption{BLEU (mean$\pm$ std) with beam size 5 for method LS or LABO on Bilingual Translation Tasks. }
    \begin{center}\small
    \begin{tabular}{lcccccc}
    \toprule
         & \multicolumn{2}{c}{IWSLT'14 } & \multicolumn{2}{c}{IWSLT'14 }  & \multicolumn{2}{c}{WMT'14 }\\ \cmidrule(lr){2-3} \cmidrule(lr){4-5} \cmidrule(lr){6-7}
      Method   & (DE-EN)  & (EN-DE)& (EN-FR)  & (FR-EN) & (EN-DE)  & (DE-EN) \\ \midrule
    Transformer     &33.9\tiny $\pm$ 0.09  & 27.8\tiny $\pm$ 0.11 &40.0\tiny $\pm$ 0.15&39.0\tiny $\pm$ 0.14 & 27.1\tiny $\pm$ 0.03 & 29.8\tiny $\pm$ 0.10\\
    w/ LS     & 34.5\tiny $\pm$ 0.14 & 28.3\tiny $\pm$ 0.16&40.5\tiny $\pm$ 0.16&39.8\tiny $\pm$ 0.05 & 27.7\tiny $\pm$ 0.09 & 31.9\tiny $\pm$ 0.09\\
    w/ CP     & 34.2\tiny $\pm$ 0.15 & 27.9\tiny $\pm$  0.07& 40.4\tiny $\pm$ 0.20& 39.2\tiny $\pm$ 0.13& 27.4\tiny $\pm$ 0.13 & 30.4\tiny $\pm$ 0.08\\
    w/ FL     & 33.0\tiny $\pm$ 0.13 & 26.8\tiny $\pm$  0.16& 39.2\tiny $\pm$ 0.16& 38.3\tiny $\pm$ 0.05& 26.6\tiny $\pm$ 0.09 & 28.9\tiny $\pm$ 0.09\\
    w/ AFL     & 34.2\tiny $\pm$ 0.14 & 27.9\tiny $\pm$  0.13& 40.5\tiny $\pm$ 0.09& 39.5\tiny $\pm$ 0.16& 27.5\tiny $\pm$0.10 & 30.3\tiny $\pm$0.08\\
    \midrule
    w/ LABO (ours) & \textbf{35.2}\tiny $\pm$ 0.07 & \textbf{28.8}\tiny $\pm$ 0.15&\textbf{40.9}\tiny $\pm$ 0.08&\textbf{40.3}\tiny $\pm$ 0.05 & \textbf{28.3}\tiny $\pm$ 0.05 &\textbf{32.3}\tiny $\pm$ 0.06\\
    \bottomrule
    \end{tabular}
    \end{center}
    \label{tab:MT}
\end{table*}
\section{Experiments}

We evaluate the performance of our proposed LABO method on both Machine Translation and Computer Vision. For machine translation, we evaluate on the IWSLT'14~\cite{cettolo2014report}, IWLST'17~\cite{cettolo-etal-2017-overview} and WMT'14~\cite{wmt} datasets using transformer-base models and for image classification on CIFAR-10, CIFAR-100~\citep{krizhevsky2009learning} and ImageNet2012~\citep{DBLP:conf/cvpr/DengDSLL009} using ResNet-based models of various sizes (parameters). All experiments were performed on one or more NVIDIA Tesla (V100) GPUs.
\subsection{Experiments on Machine Translation}
We evaluate our method on six machine translation benchmarks including IWSLT'14 German to English (DE-EN), English to German (EN-DE), English to French (EN-FR), French to English (FR-EN), WMT'14 English to German (EN-DE) and German to English (De-EN) benchmark. 
We use the 6-layer encoder-decoder transformer as our backbone model for all experiments. We follow the hyper-parameter settings for the architecture and training reported in~\cite{DBLP:conf/icml/GehringAGYD17, DBLP:conf/nips/VaswaniSPUJGKP17}. Specifically, we train the model with a maximum of 4,096 tokens per mini-batch for 150 or 50 epochs on IWSLT and WMT datasets, respectively. For optimization, we apply Adam optimizer with $\beta_1=0.9$ and $\beta_2=0.98$ and weight decay 1e-4. For LABO, we also perform explore \{1.15, 1.25\} for the only hyper-parameter $\tau = \beta/\alpha$. We report on the BLEU-4~\cite{papineni2002bleu} metric to compare between the different models\footnote{Our experiments were conducted with Fairseq toolkit (github.com/pytorch/fairseq).}.

\paragraph{Baselines.} We compare our methods with three baselines that try to handle the over-confidence problem. LS uses the combination of a one-hot vector and a uniform distribution to construct the target distribution. CP~\cite{DBLP:conf/iclr/PereyraTCKH17} punishes over-confidence by regularizing the entropy of model predictions. FL~\cite{focal_loss} utilizes the focal loss to assign smaller weights to well-learned tokens for every training iteration. AFL~\cite{DBLP:conf/emnlp/RaunakD0M20} is a generalize Focal loss which establish a 
the trade-off for penalizing low-confidence predictions.

\paragraph{Bilingual Translation.} Tab.~\ref{tab:MT} shows the results of IWLST'14 DE-EN, EN-DE, EN-FR, FR-EN and WMT'14 EN-DE, DE-EN translation tasks. The backbone transformers showed close or better BLEU scores with the numbers reported in \cite{DBLP:conf/nips/VaswaniSPUJGKP17}. All confidence penalizing techniques except FL can improve the performance of transformer models. The drop in performance of FL is consistent with the finding of long-tail phenomena in neural machine translation systems~\cite{DBLP:conf/emnlp/RaunakD0M20}. Our methods consistently outperform the baseline Transformer (w/ LS), which demonstrates its effectiveness. This is across different language pairs and dataset sizes.

\begin{table}[th!]
    \centering
    \caption{BLEU scores for method LS or LABO on multilingual Translation Tasks.}
    \begin{center}\small
    \begin{tabular}{lcc}
    \toprule
    IWSLT'17    & \multicolumn{2}{c}{BLEU}      \\
    \midrule
          Model      & DE-EN         & FR-EN         \\
                \midrule
    Transformer & 26.9          & 35.4          \\
    w/ LS       & 28.0          & 36.8          \\
    w/ LABO (ours)    & \textbf{28.4} & \textbf{37.2} \\
    \bottomrule
    \end{tabular}
    \end{center}
    \label{tab:MMT}
\end{table}
\paragraph{Multilingual Translation.} We also evaluate our LABO method on IWLST'17 (\{DE, FR\}-EN) dataset by using multilingual transformers. We learn a joint BPE code for all three languages and use sacrebleu for evaluating the test set. Tab.\ref{tab:MMT} shows LABO achieves consistent improvement over the original label smoothing on the multilingual translation dataset.

\subsection{Experiments on Image Classification}
\begin{table*}[t!]
\caption{Comparison between different Smoothed labels methods. Averaged test accuracy and training time are reported. The training time is measured on a single NVIDIA V100 GPU.}
\begin{center}\small
{
\begin{tabular}{llclclclc}
\toprule
        & \multicolumn{4}{c}{CIFAR100 (Acc. \& Time)} & \multicolumn{4}{c}{CIFAR 10 (Acc. \& Time)} \\\cmidrule(lr){2-5}\cmidrule(lr){6-9}
        & \multicolumn{2}{c}{MobileNetv2} & \multicolumn{2}{c}{ResNet18}     & \multicolumn{2}{c}{MobileNetv2} & \multicolumn{2}{c}{ResNet18}            \\\midrule
Base    & 67.98 & 1.0$\times$ &  76.92        &1.0$\times$  &         90.55  & 1.0$\times$  &  94.82     &  1.0$\times$   \\
KD     & 70.99 ($\uparrow$3.01) & 3.7$\times$ &  77.78 ($\uparrow$0.86)         &3.8$\times$    &  91.52 ($\uparrow$0.97)   & 3.7$\times$ &  \textbf{95.28} ($\uparrow$0.46)     & 3.9$\times$    \\\midrule
LS  & 68.69 ($\uparrow$0.71)    & 1.0$\times$ &    77.67 ($\uparrow$0.75)   & 1.0$\times$  &  90.82 ($\uparrow$0.27)    & 1.0$\times$ &      95.02 ($\uparrow$0.20)    & 1.0$\times$  \\
% ME      &              &               &                &                 \\\midrule
CS-KD  &  70.36 ($\uparrow$2.38) &1.1$\times$  & 77.95 ($\uparrow$1.03)     &  1.3$\times$   &  91.17 ($\uparrow$0.62)   & 1.1$\times$ &   94.90 ($\uparrow$0.08)      &  1.3$\times$  \\
TF-reg  & 70.08 ($\uparrow$2.10) &1.0$\times$ &   77.91 ($\uparrow$0.99)    & 1.2$\times$ & 90.97 ($\uparrow$0.42)    & 1.1$\times$ &   95.05 ($\uparrow$0.23)      &1.2$\times$  \\

Beta-LS &  70.45 ($\uparrow$2.47) & 1.5$\times$ & 77.83 ($\uparrow$0.91)      & 1.4$\times$   & 90.89 ($\uparrow$0.34) & 1.5$\times$  &          94.87 ($\uparrow$0.05)     &  1.6$\times$   \\
 KR-LS &  70.12 ($\uparrow$2.14) & 1.0$\times$ & 77.82 ($\uparrow$0.90) & 1.0$\times$  & 90.67 ($\uparrow$0.12) &  1.0$\times$& 94.76 ($\downarrow$0.06) & 1.0$\times$ \\
\midrule
LABO    &  \textbf{71.05} ($\uparrow$3.07) & 1.0$\times$ &{\textbf{78.10}} ($\uparrow$1.18)      & 1.0$\times$   & {\textbf{91.53}} ($\uparrow$0.98) &1.0$\times$ &    {95.21}  ($\uparrow$0.39)   &1.0$\times$  \\\bottomrule
\end{tabular}}    
\end{center}
\label{tab:cv-tf}
\end{table*}
\paragraph{Setup for CIFAR experiments.} We evaluated our method on different model architectures including MobileNetV2~\citep{DBLP:conf/cvpr/SandlerHZZC18}, and ResNet18~\citep{DBLP:conf/cvpr/HeZRS16}. 
We follow standard data augmentation schemes: random crop and horizontal flip to augment the original training images. We sampled 10$\%$ images from the training split as a validation set. The models are trained for 200 epochs with a batch size of 128. For optimization, we used stochastic gradient descent with a momentum of 0.9, and weight decay set to 5e-4. The learning rate starts at 0.1 and is then divided by 5 at epochs 60, 120 and 160.  All experiments are repeated 5 times with random initialization. For the KD experiments, we used ResNeXt29 as the teacher. All teacher models are trained from scratch and picked based on their best accuracy. To explore the best hyper-parameters, we conduct a grid search over parameter pools. We explore \{0.1, 0.2, 0.5, 0.9\} for $\alpha$ and \{5, 10, 20, 40\} for KD temperature. 
\paragraph{Results.} Next, we conduct a series of experiments on two models to compare our approach with other methods without requiring any pre-trained models on two CIFAR datasets. Note that KD is reported as a reference. All experiments are repeated 5 times with random initialization. The baseline methods include CS-KD which doesn't require training a teacher and constrains the smoothed output between different samples of the same class to be close~\citep{DBLP:conf/cvpr/YunPLS20}, TF-reg which regularizes the training with a manually designed teacher~\citep{yuan2020revisiting} and Beta-LS which leverages a similar capacity model to learn an instance-specific prior over the smoothing distribution~\citep{DBLP:conf/nips/ZhangS20}. KR-LS utilizes a class-wise target table which captures the relation of classes~\citep{DBLP:conf/aaai/DingWDSGFX21}. We follow their best hyper-parameter settings. The time for all baselines is measured based on the original implementations.~\footnote{github.com/alinlab/cs-kd}\footnote{github.com/yuanli2333/Teacher-free-Knowledge-Distillation}\footnote{github.com/ZhiluZhang123/neurips\_2020\_distillation}

Tab.~\ref{tab:cv-tf} shows the test accuracy and training time for 200 epochs of different methods on two models. It can be seen that our method consistently improves the classification performance on both lightweight and complex models, which indicates its general applicability. Besides, it shows the training time of our method is close to Base or LS methods, which show its stable efficiency over other baselines such as Beta-LS, which still requires a separate model to output a learned prior over the smoothing distribution. Our method achieves better performance than other strong baseline methods consistently. We have to mention that our computation for smoothing distribution
is deterministic and excluded from the computation graph for the gradient calculation. Therefore, the implementation of our method requires less time and space during the training as we don't need to train a separate model. 

\section{Discussion}
One of important problems of neural language models is the large discrepancy of predicted probabilities between tokens with low and high frequency, In other words, the Long-Tailed Phenomena in the neural language models~\cite{DBLP:conf/coling/ZhaoM12, DBLP:conf/acl/DemeterKD20}. In this section, we study the impact of our method on the prediction of tokens with different frequencies. 
\begin{table}[h!]
\subsection{Analysis on predictions with Low-frequency tokens}

\caption{Performance of LS and LABO on splits with different averaged token frequencies.}
\begin{center}\small
\label{tab:split}
\begin{tabular}{lccc}
\toprule
IWSLT'14 (DE-EN) & \multicolumn{3}{c}{Validation} \\
\midrule
Split            & Most     & Medium   & Least    \\
\midrule
LS               & 40.3     & 35.2     & 33.1     \\
LABO             & 40.7     & 35.8     & 33.8     \\
\
$\Delta$         & $+$0.4   & $+$0.6   & $+$0.7   \\
\midrule
IWSLT'14 (DE-EN) & \multicolumn{3}{c}{Test}       \\
\midrule
Split            & Most     & Medium   & Least    \\
\midrule
LS               & 38.4     & 33.5     & 32.1     \\
LABO             & 39.0     & 34.3     & 32.7     \\

$\Delta$         & $+$0.6   & $+$0.8   & $+$0.6 \\
\bottomrule
\end{tabular}
\end{center}
\end{table}

We first computed the averaged frequency of tokens in every source sentence $x = [x_i]|_{i=1}^N$ for the validation and test sets of IWSLT 2014 (De-En). This Frequency Score (FS) is defined in ~\cite{DBLP:conf/emnlp/RaunakD0M20}:
\begin{align}
    F(x) = \frac{\sum_{i=1}^N f(x_i)}{N},
\end{align}
where $f(x_i)$ is the frequency of the token $x_i$ in the training corpus. Next, we divide each dataset into three parts of 2400 sentences in order of decreasing FS.
Tab.~\ref{tab:split} shows the results of LS and LABO on different splits. All models perform much better on split-most than split-medium and least. Our method first demonstrates consistent improvements on three splits for both validation and test datasets.
Besides, our LABO provides at least the same magnitude of improvements on least and medium splits as the split-most.

\begin{figure}[t!]
\begin{center}
\includegraphics[width=0.45\textwidth]{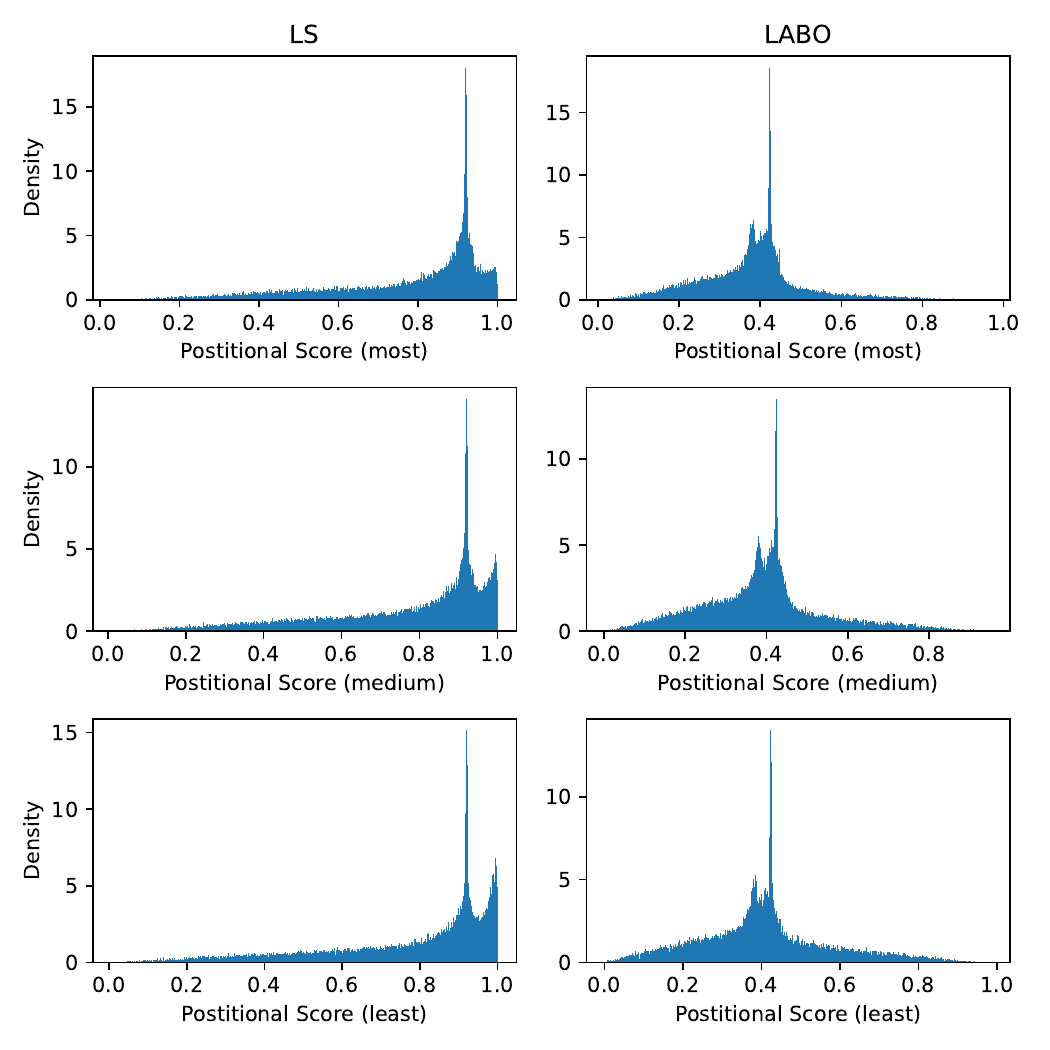} % Reduce the figure size so that it is slightly narrower than the column.
\caption{The distribution of probabilities of the selected predictions in each step of beam search for LS (left) and LABO (right) on splits with most, medium and least averaged token frequencies. All experiments are using beam size 5.}
\label{fig_analysis}
\end{center}
\end{figure}

Next, We investigate the probability distribution of the selected prediction in each step of beam search, namely, the the probability of top hypothesis finally chosen during decoding. Figure~\ref{fig_analysis}
 shows the histogram for LS and LABO on three splits. The probabilities of our LABO method concentrated on around 0.4 while the corresponding probabilities of LS concentrated on 0.9. Our method reduces the discrepancy between predicted probabilities of different tokens, which facilitates the inference process during beam search by avoiding creating extremely large probabilities. 
\section{Related Work}
There is a great body of  works inspired by the KD technique. Some of them focus on boosting performance with explicitly regularized smoothed targets. ~\citet{DBLP:conf/nips/ZhangS20} interpret student-teacher training as an amortized maximum aposteriori estimation and derive an equivalence between self-distillation and instance-dependent label smoothing. This analysis helped them to devise a regularization scheme, Beta Smoothing. However, they still use an extra model to infer a prior distribution on their smoothing technique during the training. {\citet{DBLP:conf/cvpr/YunPLS20} introduce an additional regularization to penalize the predictive output between different samples of the same class.} There are several works discussing the empirical impact of KD and giving different practical suggestions. ~\citet{tfkd_eacl} conducted extensive experiments to explore the effect of label regularization to the generalization ability of compact pre-trained language models. 

Other works propose to learn class-wise label smoothing or progressive refining of smoothed labels. {\citet{DBLP:conf/aaai/DingWDSGFX21} propose to capture the relation of classes by introducing decoupled
labels table which increases space complexity by $O(K\times K)$.} {The concurrent work~\citep{kim2020self} utilizes the model trained at $i\text{-th}$ epoch as the teacher to regularize the training at $(i+1)\text{-th}$ epoch along with annealing techniques. However, contrary to our work, it either requires a separate model as teacher, or stores the labels generated at the $i$-th epoch for training the next epoch, whereby increasing space complexity by $O(N \times K)$. ~\citet{gen_entropy} investigate the relationship between generalized entropy
regularization and label smoothing and provide empirical results on two
text generation tasks. ~~\citet{mask_ls} propose a masked Label Smoothing to remove the conflict by reallocating the smoothed probabilities based on the difference among languages.
~\citet{con-ls} propose to learn confidence for NMT by jointly training a ConNet model to estimate the confidence of the prediction.
We develop LABO motivated by a principled starting point: to generalize the smoothing distribution to a general form with neither time nor space complexity increase during the training and inference. 
}
\section{Conclusion}
Our aim in this work is to fill the accuracy gap between Label Smoothing and Knowledge Distillation techniques while maintaining the training efficiency of LS regularization. We proposed learning an instance-dependent label smoothing regularization simultaneously with training our model on the target. We began by generalizing the classical LS method and introduced our objective function by substituting the uniform distribution with a general, instance-dependent, discrete distribution. Within this  formulation, we explained the relationship between the LS and KD. Then, using a bi-level optimization approach, we obtained an approximation for the optimal smoothing function. We conducted extensive experiments to compare our model with conventional LS, KD and various state-of-the-art label regularization methods on popular MT and V benchmarks and showed the effectiveness and efficiency of our technique. Finally, we analyze the impact of our methods on the prediction of neural machine translation systems under different averaged token frequency settings and show our methods can greatly reduce the discrepancy between predicted probabilities of different tokens.  
In practice, apart from  general regularization techniques like dropout and weight decay, many advanced techniques are designed for specific tasks like FixNorm~\citep{fixnorm}, CutMix~\citep{DBLP:conf/iccv/YunHCOYC19} and SwitchOut~\citep{DBLP:conf/emnlp/WangPDN18}. We leave the future work to combine these methods with our technique. Moreover, we plan to explore the practical applications of our method for large-scale model training. One specific application is to improve the pre-training of large language models and vision transformers.
% \section*{Acknowledgments}
% We thank Mindspore,\footnote{\url{www.mindspore.cn/}} which is a new deep learning computing framework, for partial support of this work. 
\section{Limitations}
In the current work, we adapt one-step gradient descent training for the outer loop based on our bi-level optimization framework. Since this outer loop optimization doesn't have a closed-form solution, determining how many steps to perform for the outer loop for better outer optimization is still important to explore. 

\bibliography{anthology,custom}
\bibliographystyle{acl_natbib}

\appendix
\onecolumn

% \section{Appendix}

\section{Derivation of the proof for Theorem 1}

\begin{proof}
First, observe that the function $\mathcal{R}_\theta$ defined in  Eq.~\ref{eq:pseudo-kd} is a convex combination of $N$ nonnegative functions $\mathcal{R}_i = \mathbb{E}_{\hat{P}_{ls}}\left[ -\log p_\theta(j|x_i)  \right] + \beta\mathbb{KL}(P_{ls}(\cdot|x_i) \Vert U(\cdot))$, for $i = 1, \dots, N$. We will show that each of $\mathcal{R}_i$ is a convex function of the components of simplex $P_{ls}(\cdot \vert x_i)$ by computing the Hessian matrix of $\mathcal{R}_i$
with respect to   $P_{ls}(\cdot \vert x_i)$:
\begin{align}
    \mathbf{H}(R_i(P_{ls}(1), \cdots, P_{ls}(K))&=\begin{bmatrix}
\frac{\partial^2 \mathcal{R}_i}{\partial P_{ls}(1)^2}, & \frac{\partial^2 \mathcal{R}_i}{\partial P_{ls}(1)\partial P_{ls}(2)}, & \cdots, & \frac{\partial^2 \mathcal{R}_i}{\partial P_{ls}(1)\partial P_{ls}(K)} \\ 
 \frac{\partial^2 \mathcal{R}_i}{\partial P_{ls}(2)\partial P_{ls}(1)},&  \frac{\partial^2 \mathcal{R}_i}{\partial P_{ls}(2)^2}, & \cdots, & \frac{\partial^2 \mathcal{R}_i}{\partial P_{ls}(2)\partial P_{ls}(K)}\\ 
 \vdots, & \vdots,& \ddots,&\vdots \\ 
\frac{\partial^2 \mathcal{R}_i}{\partial P_{ls}(K)\partial P_{ls}(1)}, & \frac{\partial^2 \mathcal{R}_i}{\partial P_{ls}(K)\partial P_{ls}(2)},  &\cdots, &\frac{\partial^2 \mathcal{R}_i}{\partial P_{ls}(K)^2} \\
\end{bmatrix}\\
% &=\begin{bmatrix}
% \frac{\beta }{P_{ls}(1)}, & 0, & 0,  & \cdots, & 0 \\ 
%  0,&  \frac{\beta }{P_{ls}(2)}, & 0, & \cdots, & 0 \\ 
%  0, & 0, & \ddots, & \cdots,& 0 \\ 
% \vdots, & \vdots, & \vdots, &\ddots, &\vdots \\
% 0, & 0, & \cdots, &0, &\frac{\beta }{P_{ls}(K)} \\
% \end{bmatrix}\\
&=\text{diag}(\frac{\beta}{P_{ls}(1)}, \frac{\beta}{P_{ls}(2)}, \cdots, \frac{\beta}{P_{ls}(K)})\nonumber
\end{align}
When $\beta$ is greater than zero, the Hessian is positive definite. Therefore, each $R_i$ is a convex function of the components of $P_{ls}(\cdot\vert x_i)$. As a result, $\mathcal{R}_\theta$ is a convex function of the collection of components of every simplex $P_{ls}(\cdot \vert x_i)$. 

For simplicity, we derive the global optimum solution for each $\mathcal{R}_i$ with a Lagrangian multiplier:
\begin{align}
    L_i(P_{ls}, \lambda_L) = \sum_{j'=1}^K\left[ -\hat{P}_{ls}(j')\log p(j')  + \beta\cdot 
P_{ls}(j')\log\frac{P_{ls}(j')}{1/K} \right] + \lambda_L(\sum_{j'=1}^K P_{ls}(j') -1),
\end{align}
where we omit the dependency on $x_i$ to simplify the notation.
We set the corresponding gradients equal to $0$ to obtain the global optimum for $j = 1, \dots, K$.
\begin{align}
    \frac{\partial L_i}{\partial P_{ls}(j)} =& -\alpha \log p(j) + \beta\cdot \log P_{ls}(j) + \beta + \beta \log K  + \lambda_L = 0 \\\nonumber
      P^*_{ls}(j) =& \exp(\frac{\alpha}{\beta}\log p(j)) \cdot \exp(\frac{-\beta -\beta \log K - \lambda_L}{\beta })\\
      =& \exp(\frac{\alpha}{\beta}\log p(j)) \cdot C_{ls}
\end{align}
Since $\sum_{j'=1}^K P_{ls}(j') = 1$, we have
\begin{align}
    \sum_{j'=1}^K P_{ls}(j') = & \sum_{j'=1}^K \exp(\frac{\alpha}{\beta}\log p(j')) \cdot \exp(\frac{-\beta -\beta \log K - \lambda_L}{\beta })=1 \\
    C_{ls} =  & \exp(\frac{-\beta -\beta \log K - \lambda_L}{\beta }) = \frac{1}{ \sum_{j'=1}^K \exp(\frac{\alpha}{\beta}\log p(j'))}
\end{align}
So the optimal $ P_{ls}^*(j)$ is given by the formula:
\begin{align}
    P_{ls}^*(j) = \frac{\exp(\frac{\alpha}{\beta}\log p(j))}{\sum_{j'=1}^K\exp(\frac{\alpha}{\beta}\log p(j')) }  = \frac{p(j)^{\alpha/\beta}}{\sum_{j'=1}^Kp(j')^{\alpha/\beta}}.
\end{align}
% We can compute the Hessian matrix of Eq.~\ref{eq:pseudo-kd} w.r.t. $P_{ls}$:
% \begin{align}
%     \mathbf{H}(R(P_{ls}))&=\begin{bmatrix}
% \frac{\partial^2 L}{\partial P_{ls}(1)^2}, & \frac{\partial^2 L}{\partial P_{ls}(1)\partial P_{ls}(2)}, & \cdots, & \frac{\partial^2 L}{\partial P_{ls}(1)\partial P_{ls}(K)} \\ 
%  \frac{\partial^2 L}{\partial P_{ls}(2)\partial P_{ls}(1)},&  \frac{\partial^2 L}{\partial P_{ls}(2)^2}, & \cdots, & \frac{\partial^2 L}{\partial P_{ls}(2)\partial P_{ls}(K)}\\ 
%  \vdots, & \vdots,& \ddots,&\vdots \\ 
% \frac{\partial^2 L}{\partial P_{ls}(3)\partial P_{ls}(1)}, & \frac{\partial^2 L}{\partial P_{ls}(3)\partial P_{ls}(2)},  &\cdots, &\frac{\partial^2 L}{\partial P_{ls}(K)^2} \\
% \end{bmatrix}\\
% % &=\begin{bmatrix}
% % \frac{\beta }{P_{ls}(1)}, & 0, & 0,  & \cdots, & 0 \\ 
% %  0,&  \frac{\beta }{P_{ls}(2)}, & 0, & \cdots, & 0 \\ 
% %  0, & 0, & \ddots, & \cdots,& 0 \\ 
% % \vdots, & \vdots, & \vdots, &\ddots, &\vdots \\
% % 0, & 0, & \cdots, &0, &\frac{\beta }{P_{ls}(K)} \\
% % \end{bmatrix}\\
% &=\text{diag}(\frac{\beta}{P_{ls}(1)}, \frac{\beta}{P_{ls}(2)}, \cdots, \frac{\beta}{P_{ls}(K)})\nonumber
% \end{align}
% When $\beta$ is greater than zero, the Hessian is positive definite. So Eq.~\ref{eq:pseudo-kd} achieves the minimum at $ P_{ls}^*(\cdot)$.
% We could guarantee that the  $R(P_{ls})$ is convex w.r.t. $P_T$ when we set $\beta < \alpha$ and $0<P_T(j)<1$ for $j \in [1,\dots, K]$, as the Hessian matrix $H$ is a positive definite. Therefore, we get the  minimum value on the global optimum.
\end{proof}

\section{Derivation from optimal smoothing to softmax output with temperature}
When $\tau = \frac{\beta}{\alpha}$, we could have 
\begin{align}
    P^*_{ls}(c\vert x_i) &= \frac{p_\theta(c\vert x_i)^{\frac{1}{\tau}}}{\sum_{j}p_\theta(j|x_i)^{\frac{1}{\tau}}} 
    =\frac{(\frac{e^{z_{i,c}}}{\sum_m e^{z_{i,m}}})^{\frac{1}{\tau}}}{\sum_j(\frac{e^{z_{i,j}}}{\sum_m e^{z_{i,m}}})^{\frac{1}{\tau}}} \notag \\
    &= \frac{(e^{z_{i,c}})^{\frac{1}{\tau}}}{\sum_j(e^{z_{i,j}})^{\frac{1}{\tau}}} = \frac{e^{\frac{z_{i,c}}{\tau}}}{\sum_j e^{\frac{z_{i,j}}{\tau}}}
\end{align}
% \section{Appendix}
\label{sec:appendix}
\twocolumn
\section{Additional Results on ImageNet}
\paragraph{Setup for ImageNet experiments.} We evaluated our method on two model architectures including ResNet50 and ResNet152 with standard data augmentation schemes including random resize cropping and random horizontal flip. The models are trained for 90 epochs with a batch size of 256. We use SGD for optimization with a momentum of 0.9, and weight decay set to 1e-4. The learning rate starts at 0.1 and is then divided by 10 at epochs 30, 60 and 80. 
\begin{table}[th!]
\caption{Comparison between different Smoothed labels methods. Validation accuracy and training time are reported. The training time is measured on 4 NVIDIA V100 GPUs.}
\begin{center}\small
\begin{tabular}{llclc}
\toprule
        & \multicolumn{4}{c}{ImageNet (Acc. \& Time)}  \\
        \cmidrule(lr){2-5}
        &  \multicolumn{2}{c}{ResNet50}     & \multicolumn{2}{c}{ResNet152}          \\\midrule
Base    &   75.81         & 1.0$\times$ &       77.92      & 1.0$\times$    \\
LS  &     76.17 ($\uparrow$0.36)   & 1.0$\times$ &    78.33 ($\uparrow$0.41)   & 1.0$\times$ \\
% ME      &              &               &                &                 \\\midrule
TF-reg  & 76.21 ($\uparrow$0.40)    & 1.1$\times$ &     78.12 ($\uparrow$0.20)   & 1.1$\times$\\
Beta-LS &  76.13 ($\uparrow$0.32)    & 1.5$\times$   &      78.56 ($\uparrow$0.64)  & 1.6$\times$  \\
 KR-LS &  76.32 ($\uparrow$0.51) & 1.3$\times$ & 78.48 ($\uparrow$0.56) &  1.3$\times$ \\
\midrule
LABO    &  {\textbf{76.55}} ($\uparrow$0.74)      &  1.1$\times$  &    {\textbf{78.62}} ($\uparrow$0.70)     &  1.1$\times$ \\\bottomrule
\end{tabular}
\end{center}
\label{tab:imagenet}
\end{table}
\paragraph{Results.} Tab.~\ref{tab:imagenet} shows the accuracy and training time for one epoch of different methods on two models. First, our method can consistently improve the classification performance, which indicates its robustness on the large-scale dataset. Again, our method achieves better performance with compared with other smoothing functions with a moderate training time increase over Base. Our computation overhead for smoothing distribution is introduced by computing Eq.~\ref{theorem1} which
is deterministic and excluded from the computation
graph for the gradient calculation, hence, our method is more efficient than other latest advanced LS techniques.

\section{Data Statistics}
\begin{table}[h!]
\begin{center}\small

\begin{tabular}{lrrr}
\toprule
Dataset          & {Train}  & Validation                 & Test     \\
\midrule
IWSLT'14 (DE-EN) & 160,239 & 7,283                       & 6,750                        \\
IWSLT'14 (FR-EN) & 168,151                     & 7,643                       & 4,493                        \\
WMT'14 (EN-DE)   & 3,900,502                    & 39,414                      & 3,003                        \\
IWSLT'17 (DE-EN) & {209,522} & 7,887                       & 5,670                        \\
IWSLT'17 (FR-EN) & {236,652} & 8,277                       & 7,275                        \\
CIFAR10          & 45,000                      & 5000                       & 10,000                       \\
CIFAR100         & 45,000                      & 5,000                       & 10,000                       \\
ImageNet         & 1,281,167                  & {50,000} & {100,000}\\
\bottomrule
\end{tabular}
    
\end{center}
\end{table}
% This is an appendix.

\section{Experimental Details for MT}
We conduct experiments by using the same hyper-parameters for fair comparisons. Before training, we first apply BPE~\cite{bpe} to tokenize the corpus for each language pair.
During the training, we set the label smoothing parameter to 0.1. We follow previous work to use Adam optimizer with betas
to be (0.9,0.98) and the learning rate is 7e-4 for WMT and 5e-4 for the rest of tasks. During warming-up steps, the initial learning rate is 1e-7 and there are 1000 warm-up steps. For the warm-up steps of our smoothing, we use 10000 for WMT and 5000 for other tasks. Dropout rate is set to 0.3 and weight decay is set to 0.0001 for all
experiments. We pick the checkpoint with the best performance on the validation set before inferring on the test set with beam size 5. 
\end{document}